\documentclass[conference]{IEEEtran}
\IEEEoverridecommandlockouts

\usepackage{cite}
\usepackage{amsmath,amssymb,amsfonts}
\usepackage{algorithmic}
\usepackage{graphicx}
\usepackage[hidelinks]{hyperref}
\usepackage{hyperref}

\usepackage{algorithm}
\usepackage{algorithmic}

\usepackage{tabularx}
\usepackage{multirow}
\usepackage{float}
\usepackage{amssymb}
\usepackage{makecell}
\usepackage{booktabs} 
\usepackage{subcaption}
\usepackage{textcomp}
\usepackage{xcolor}
\def\BibTeX{{\rm B\kern-.05em{\sc i\kern-.025em b}\kern-.08em
    T\kern-.1667em\lower.7ex\hbox{E}\kern-.125emX}}
\begin{document}

\title{XKV: Personalized KV Cache Memory Reduction for Long-Context LLM Inference\\
}
 \author{\IEEEauthorblockN{Weizhuo Li\IEEEauthorrefmark{2}, Zhigang Wang\IEEEauthorrefmark{3}, Yu Gu\IEEEauthorrefmark{2}, Ge Yu\IEEEauthorrefmark{2}\\
 \IEEEauthorblockA{\IEEEauthorrefmark{2}Northeastern University, \IEEEauthorrefmark{3}Guangzhou University} 
 \IEEEauthorblockA{lwzzzz@foxmail.com, wangzhiganglab@gmail.com, guyu@mail.neu.edu.cn, yuge@mail.neu.edu.cn}
}}

\maketitle

\begin{abstract}
Recently the generative Large Language Model (LLM) has achieved remarkable success in numerous applications. Notably its inference generates output tokens one-by-one, leading to many redundant computations. The widely-used KV-Cache framework makes a compromise between time and space complexities. However, caching data generates the increasingly growing memory demand, that can quickly exhaust the limited memory capacity of the modern accelerator like GPUs, particularly in long-context inference tasks. Existing studies reduce memory consumption by evicting some of cached data that have less important impact on inference accuracy. But the benefit in practice is far from ideal due to the static cache allocation across different LLM network layers.

This paper observes that the layer-specific cached data have very different impacts on accuracy. We quantify this difference, and give experimental and theoretical validation. We accordingly make a formal analysis and shows that customizing the cache size for each layer in a personalized manner can yield a significant memory reduction, while still providing comparable accuracy. We simulate the cache allocation as a combinatorial optimization problem and give a globally optimal solution. In particular, we devise a mini- and sampling-based inference over a lightweight variant of the LLM model, so as to quickly capture the difference and then feed it into the personalized algorithms. Extensive experiments on real-world datasets demonstrate that our proposals can reduce KV cache memory consumption by 61.6\% on average, improve computational efficiency by 2.1$\times$ and then increase the throughput by up to 5.2$\times$.
\end{abstract}


\section{Introduction}
Since GPT-3, hundreds of Large Language Models (LLMs) have been released\cite{achiam2023gpt4}. The workflow of nowadays LLM services roughly consists of two stages, i.e., training the model and subsequently performing application-specific inference tasks. Training requires extremely huge resources because Forward Propagation (FP) and Backward Propagation (BP) are alternatively run multiple times over millions of training samples. They are typically performed on dedicated AI clusters with thousands of GPU accelerators, built by only a few Top IT companies, like RSC in Meta\footnote{https://blogs.nvidia.com/blog/meta-ai-supercomputer-dgx/}. By contrast, inference is lightweight since a task only runs FP once to process a single input sample. As a result, LLM inference has been widely explored in many downstream applications, including dialogue systems\cite{chiang2023vicunaduihua}, code completion\cite{roziere2023code}, document summarization\cite{Fabbri_Li_She_Li_Radev_2019summar}, and language translation\cite{yang2023bigtranslate}. Among them, most are transformer-based\cite{vaswani2017attention} generative inference tasks where new tokens are continuously generated based on the original input context and all of already output tokens. Such a behavior is computation- or memory-intensive. That still poses significant challenges for efficient inference in practice, due to the limited hardware resources in modest companies and academia.

A transformer architecture\cite{vaswani2017attention} typically consists of tens or even hundreds of stacked network layers in generative LLMs. During FP in inference, a sample as context is first decomposed into a series of input tokens, and the latter are then fed into the model to extract features layer-by-layer. The finally output token will be appended into the tail of the original input sequence, so as to infer the next token. This inference pass is continuously run until the length of all outputs is greater than a pre-set upper bound. 
The core operation in a transformer layer is computing the self-attention score using Eq.~\eqref{equ:qkv}. The $Q$, $K$ and $V$ vectors can capture the relationship among all input tokens to guarantee the generalization and accuracy in inference. The three vectors are calculated in each inference pass. Recall that an inference task always concatenates already existing old tokens with the newly output one, and then feeds them as a whole into the first layer. Existing studies have stated that in the $K$ and $V$ vectors, elements associated with old tokens are repeatedly calculated{\cite{radford2019languagekvcachetichu}}. Caching these K-V pairs is a prominent solution to eliminate redundant computations\cite{radford2019languagekvcachetichu}, that has been widely employed in mainstream LLM services.
\begin{equation}\label{equ:qkv}
Z=Softmax \left( \frac{QK^T}{\sqrt{d_k}} \right) V
\end{equation}

\begin{figure}
    \centering
    \includegraphics[width=1\linewidth]{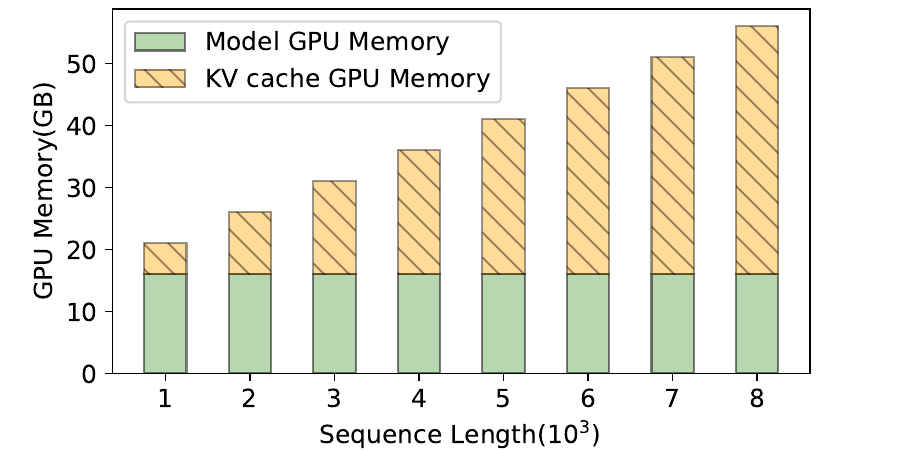}
    \caption{Comparison on memory usage respectively for model parameters and KV cache, when running inference tasks with the batch size of 10, on LLama-3.1 with 8 billion parameters. The sequence length of each input varies from 1$\times\!10^3$ to 8$\times\!10^3$.}
    \label{fig:11}
\end{figure}

Notably the dimension of K-V vectors is proportional to the length of concatenated input. As the inference proceeds, the memory consumed by cached data significantly increases. In fact, under the caching solution, the generative LLM inference is transferred from computation-intensive into memory-intensive. Fig.~\ref{fig:11} plots memory consumption versus the input sequence length, using the LLama-3.1\cite{dubey2024llama3} model as an example. We clearly see that the memory consumption of parameters is fixed as 16GB. While, with the increase of sequence length, the consumption of K-V cache increases from 5GB up to 40GB, which even dominates the overall memory requirement. Clearly, there is an imperative need for memory-efficient generative LLM inference.
A straightforward solution is to employ more devices{\cite{298687distserve},\cite{agrawal2024taming},\cite{lin2024infinite}}, but at expense of huge hardware investments and inevitable cooperation delay. Thus, most efforts focus on compromising inference accuracy for memory reduction. Some of them attempt to merge similar elements in cache and then reuse the merged values in inference\cite{zhang2024cam,wang2024KVMerger}. Another research line stores cached data with few bits (quantization) and recovers them when performing computations\cite{kang2024gear,zhang2024pqcache}. Recently many tests validate that tokens in the input sequence have different impacts on the accuracy of the newly generated token{\cite{xiao2024streamingllm,zhang2024h2o}}. This motivates researchers to evict K-V pairs w.r.t. less important tokens to reduce the memory consumption\cite{li2024snapkv,yang-etal-2024-pyramidinfer}. The three optimizations are compatible with each other, and this paper primarily concerns on eviction. 

More specifically, an inference task needs to first prefill $K$-$Q$-$V$ vectors w.r.t. the original input sequence, and then continuously generate output tokens one by one\cite{298687distserve}. Conventionally, the importance of cached data is evaluated by the attention score between old tokens and the new output one. A majority of existing solutions thereby cannot evict any data during the prefill stage, since now no output is available\cite{NEURIPS2023_aScissorhands,ge2024fastgen}. They work well in the traditional short-context scenario. However, the requirement issued by end-users is becoming complex. The length of the original input sequence (context) is usually greater than that of the output sequence. Taking the long-dependency question-answering task as an example, the context has 40,000 tokens on average, while the output contains only 500 tokens\cite{li2023loogle}. That significantly challenges the memory burden in prefill. The inference task might immediately crash without any output, due to the out-of-memory error. The most recent works logically regard the last few tokens in the context as virtual outputs to distinguish the importance for prefill eviction\cite{li2024snapkv,yang-etal-2024-pyramidinfer,zhang2024pyramidkv}. However, the attention score shown in Eq.~\eqref{equ:qkv} is layer-specific. These pioneering works ignore the variation of importance across layers. The memory reduction is then still far from ideal.

We are aware that the heavy conflict between huge memory requirements in LLMs and limited hardware resources is coming with strong and ever-growing demands on efficient data storage, which is a traditional yet important research topic in the database (DB) community. Till now, many attentions have been attracted but they mainly focus on the model training\cite{db1},\cite{db2},\cite{db3},\cite{db4}. Few of them can solve the memory reduction issue in the prefill stage. This paper as the first attempt fills this research gap.

We systematically perform a fine-grained exploration about the \underline{D}ynamic \underline{D}ifferences of \underline{I}mportance \underline{D}istribution (DDID) phenomenon. We present a novel $R$ function to quantify the key information embedded in preserved cached data. Extensive pioneering tests then validate the DDID insight. That is, when performing eviction due to the memory capacity constraint, the impacts of cached K-V pairs on accuracy vary with layers. We next establish the relationship between layer-specific $R_i$ and the final inference accuracy. By a theoretical analysis, we find that the accuracy guarantee is equivalent to the maintenance of $R_i$. Motivated by this, we build a memory-efficient inference algorithm where the cache size specific to each layer can be dynamically tuned, based on the hierarchy of the model network and the inference application. That achieves prominent memory reduction while still providing with comparable and even better accuracy.

Given the accuracy constraint, we need to carefully customize the memory allocation among different layer-specific caches, so as to maximize the memory savings. However, solving this combinatorial optimization problem is time-consuming. Benefiting from the equivalence between $R_i$ and the final accuracy, we give a greedy-based optimization rule to quickly determine the allocation layer-by-layer. Besides, the allocation across layers depends on statistics of DDID, but this knowledge is not available before launching any inference tasks. We tackle this nut by a mini-prefill procedure where several tasks are run on a lightweight variant of the normal LLM model to collect attention-related statistics. Here we do not cache any data to avoid possible out-of-memory errors. Nevertheless, running mini-prefill and algorithms is still not free. We investigate the allocation results of different inference tasks and find that they have the similar distribution across layers. Thus, given a specific application, we can sample some tasks in advance and determine the allocation with their average, instead of repeatedly running mini-prefill and algorithms for every inference task.

We summarize our contributions below.
\begin{itemize}
    \item Presenting the DDID insight and establishing the equivalence between preserved information in each layer's cache and the final inference accuracy, with experimental and theoretical validation. The layer-specific cache size thereby can be customized to reduce memory consumption with acceptable accuracy loss (Sec.~\ref{research}).
    \item Proposing a greedy-based optimization rule to quickly customize cache allocation based on DDID statics collected by a lightweight LLM model. The sampling technique further reduces the runtime delay caused by running algorithms and collecting statistics (Sec.~\ref{sec:impl}). 
    \item Performing extensive experiments on real datasets against up-to-date competitors (Sec.~\ref{sec:exp}). The KV cache memory usage is reduced by 61.6\% on average. The computational efficiency is improved by 2.1$\times$. The throughput thereby increases by up to 5.2$\times$.
\end{itemize}

\section{Related Works}\label{sec:rw}
There are a flurry of efforts devoted into reducing the memory requirement of KV cache. By our investigations, they basically fall into three categories, merging\&reusing, quantization, and eviction. Below we overview them respectively to highlight our contributions.

\subsection{Merging\&Reusing}
This research line aims to identify similar data in cache and then merge them for memory reduction. The merged data is decoded for reusing during inference. Researchers present many policies to reduce the merging-reusing error, so as to mitigate the accuracy loss. In particular, Cache Merging (CaM)\cite{zhang2024cam} does not directly evict unimportant cached data. Instead, it merges such data with them normally preserved in the cache. KVMerger\cite{wang2024KVMerger} identifies candidate mergeable KV cache pairs using a Gaussian kernel-weighted technique, which is especially tailored for long-context tasks. Different from them, YOCO\cite{sun2024yoco} gives a new decoder-decoder architecture to enable an across-layer merging. That is, given an output token, YOCO merges its associated Key-Value pairs distributed among layer-specific caches. When inferring the subsequent tokens, it employs a self-decoder to recover data required at different layers, based on the single merged pair.

\subsection{Quantization}

The main idea behind quantization is to convert high-precision data in the KV cache into low-precision representations. Several advanced methods have been proposed to effectively quantize data with nearly-zero accuracy penalty.

GEAR\cite{kang2024gear} applies the ultra-low precision quantization to a majority of cached data with roughly equal-values. The high compression ratio significantly reduces the peak memory usage. To address the resulting quantization error, GEAR uses a low-rank matrix for approximation and a sparse matrix for outlier entries. Further, KVQuant \cite{hooper2024kvquant} notices that the value distribution of cached data is channel- and vector-specific. It thereby gives four quantization policies to cope with different scenarios, to strike a fine-grained balance between memory reduction and inference accuracy. KIVI\cite{liu2024kivi} goes further by separating cached data. It quantizes the K-cache by channel and the V-cache by token. FlexGen\cite{sheng2023flexgen}, on the other hand, focuses on the hardware-aware quantization, including GPU memory, CPU memory, and out-of-core disks. It thereby significantly improves the overall throughput. Lastly, PQcache \cite{zhang2024pqcache} models the KV-cache management as a classic embedding retrieval problem in the database community. It then uses Product Quantization (PQ) to perform memory efficient inference with low service latency and high accuracy. That opens up a new field of optimizing the KV cache costs.

\subsection{Eviction}
Many pioneering studies have already revealed that the cached KV pairs have very different importance in inference. Inspired by this, a straightforward optimization is to preserve crucial pairs while evicting other less important ones. Clearly, the key issue for eviction is how to effectively distinguish importance. Recall that an inference task consists of prefill data based on input and generating output tokens. Existing efforts thereby perform identification and eviction respectively on the two stages. 

Most of works focus on the generation stage. H2O\cite{zhang2024h2o} dynamically filters the unimportant pairs by the real-time attention scoring function. StreamingLLM\cite{xiao2024streamingllm} always preserves the attention sinks and caches the KV pairs of the most recently generated tokens. That significantly reduces the memory consumption but at expense of heavy information loss. Scissorhands\cite{NEURIPS2023_aScissorhands} assumes that the set of tokens currently having high attention score can keep steady in the following generation. It thereby always cache their KV pairs without complex replacement behaviors. FastGen\cite{ge2024fastgen} identifies attention modules to dynamically evict pairs. ADORE\cite{zhang-etal-2024-adore} can reconstruct already evicted data if they are crucial for the high-quality generation. ALISA\cite{zhao2024alisa} devises a Sparse Window Attention algorithm to distinguish the importance of KV pairs for the downstream generation. It accordingly prioritizes data for smart eviction. EasyKV\cite{guo2024attentioneazykv} measures the importance based on not only attention score but also the $\ell$1 norm of value vectors. In general, all of these methods identify important KV pairs by analyzing the relationship between the newly generated token and already existing tokens. They cannot evict data and hence reduce memory consumption until the inference task completes prefill and then proceeds into the generation stage. However, currently, the growing length of the inference input significantly increases the volume of prefilled data. The prefill stage even dominates the peak memory consumption during inference. That possibly renders inference failure in advance in the prefill stage, due to the out-of-memory error. 

Recent works thereby study how to directly evict data in the prefill stage. Now the new output token is not available. The main idea is to regard the last few tokens in the input sequence as the alternative, since they usually play important roles in generating the new output token. Afterwards, the attention score can be normally analyzed as done in generation-based solutions. SnapKV\cite{li2024snapkv} follows this design and uses an observation window to quantify the number of tokens regarded as the alternative. PyramidInfer\cite{yang-etal-2024-pyramidinfer} and PyramidKV\cite{zhang2024pyramidkv} further capture the cross-layer differences about the importance of tokens. They group hierarchical layers in a coarse-grained manner and then pre-set the cache size for each group, following a pyramid-shaped distribution. Notably the shape of the pyramid is determined by static hyperparameters that are calculated by features across coarsen-grained groups.
However, by our investigation, such an important difference dynamically evolves with not only space (among hierarchical layers) but also time (inference task). Thus, the memory reduction of PyramidInfer and PyramidKV is still far from ideal. Besides, Ada-KV\cite{feng2024adakvoptimizingkvcache} pays attention to the differences of attention heads and accordingly pre-sets the cache sizes for them. It can work together with the layer-centric works\cite{yang-etal-2024-pyramidinfer,zhang2024pyramidkv}.

In summary, the three kinds of memory reduction techniques are compatible with each other. Among them, eviction as a representative has been extensively studied. However, few of related works focus on the pre-filling stage that usually dominates the peak memory consumption in current long-context tasks. Our new spatial-temporal-aware optimization can capture the fine-grained importance difference among tokens. Accordingly, it customizes the cache size for each layer and dynamically tunes the setting. That is technically orthogonal to the existing focus.

\section{A Dynamic KV Cache Allocation across Hierarchical Layers}\label{research}

\begin{table}
 \caption{Notations}\label{t1}
  \centering
 \resizebox{0.48\textwidth}{!}{
\begin{tabular}{ll}
\toprule
Symbol&Explanation \\
\midrule
$Q$& The query matrix obtained by linear transformation of the input\\
$K$& The key matrix obtained by linear transformation of the input\\
$V$& The value matrix obtained by linear transformation of the input\\
$d_k$& The dimension of the key matrix\\
$Z$& The output after self - attention\\
$W_{attn}$& The calculated attention weight matrix\\
$ows$& The size of observation window\\
$R$&The ratio of importance retention\\
$w$&The attention score distribution vector\\
$n$&The KV cache size of one layer\\
$ISR$&The ratio of importance retention to size\\
\bottomrule
\end{tabular}
  }
\end{table}
In this section, we conduct detailed experimental studies and theoretical analyses on DDID in the hierarchical structure during the LLM inference process. Finally, we will describe the selection of personalized KV cache reduction strategy as an optimization problem related to DDID. \autoref{t1}
 summarizes the main notations through this paper, for better understanding.
 
\subsection{DDID: Dynamic Differences of Importance Distribution}
The core idea of KV cache reduction strategy based on eviction is to retain KV pairs corresponding to tokens that are relatively more important for the current inference. During the prefilling stage, the LLM processes the entire long-context input to infer the first new token. Therefore, at this stage, the core idea is to retain KV pairs corresponding to tokens in the input sequence that are crucial for this new token. Existing studies have validated that tokens important for current generation are also important in subsequent generation\cite{NEURIPS2023_aScissorhands}. Thus, KV pairs retained during the prefilling stage continuously play a critical role during the entire inference computation. Recall that the most significant challenge in eviction is how to evaluate the importance of tokens for inference. In most existing studies\cite{xiao2024streamingllm,zhang2024h2o}, this token importance is determined based on the intermediate results when computing the the attention weight in their self-attention mechanism. The attention weight can capture the correlation among all tokens, that can be computed by Eq.~\eqref{Wattn}.
\begin{equation}\label{Wattn}
W_{attn}=softmax(\frac{QK^T}{\sqrt{d_k}})
\end{equation}

\begin{figure}
    \centering
    \includegraphics[width=1\linewidth]{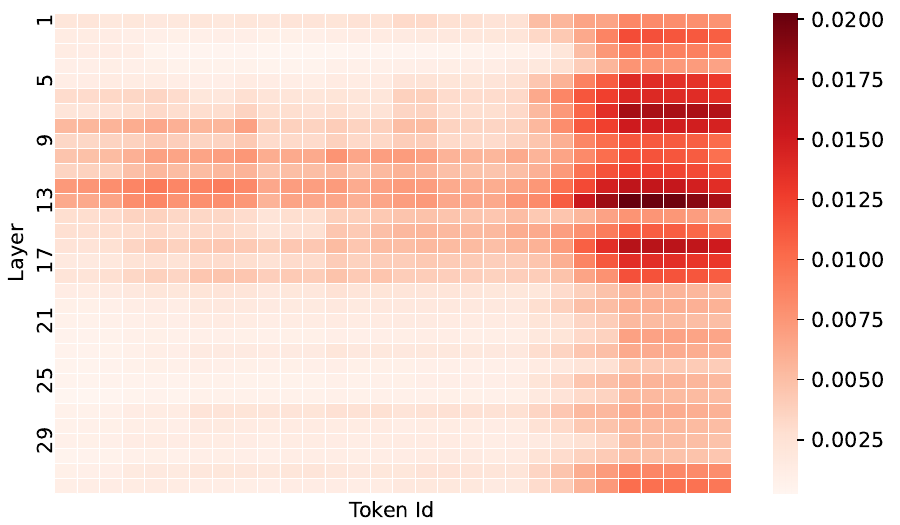}
    \caption{The heatmap of the attention score}
    \label{fig:hot}
\end{figure}

Different from the generation stage, $W_{attn}$ calculated during the prefilling stage indicates the correlation among all tokens in the original input sequence, instead of the generated output token Consequently, it is not reasonable to directly utilize the $W_{attn}$-based importance of tokens in the input sequence to infer the first new token.
However, several studies \cite{li2024snapkv, yang-etal-2024-pyramidinfer, zhang2024pyramidkv, liu2024lost} have demonstrated that the last few tokens in a long-context input sequence are critical for inferring the first new token during the prefilling stage. This insight helps to crack this difficult nut. These last few tokens falling into a pre-defined observation window\cite{li2024snapkv, yang-etal-2024-pyramidinfer, zhang2024pyramidkv}, play a pivotal role in generating the first new token. We also use the observation window as a reference, and reformulate the problem. That is, instead of measuring the importance of all tokens in the input sequence, we just assess the correlation between tokens outside the observation window and those within it. As shown in Eq.~\eqref{W}, we can obtain the distribution vector $w$ of the attention score. It explicitly measures the importance of the tokens in the input sequence for inferring the first token in the prefilling stage, by a specific function $f$ (described later).
\begin{equation}\label{W}
w=f\left( W_{attn},\, ows \right) 
\end{equation}
Here, $W_{attn}$ is attention weight matrix, $ows$ is the observation window size. The higher the score corresponding to a token in $w$ is, the more important the token is to the observation window. Meanwhile, it is also more important for inferring the first new token. This serves as an important guideline for whether to evict or retain KV pairs of a token during the prefilling stage. Due to the multi-layer structure of LLMs, the input at each layer dynamically changes during the inference process. The intermediate computation result $W_{attn}$ accordingly varies. As a result, $w$ derived from $W_attn$ through $f$ thereby dynamically varies across layers. This cross-layer variation leads to differences in the importance distributions. That can be used to guide the eviction strategy at each layer. To validate such dynamic differences, we conduct a series of experiments on a 32-layer open-source LLM and plot this phenomenon for better understanding. In particular, during the prefilling stage of inference, we extract $w$ from each layer and then visualize part of the results as shown in \autoref{fig:hot}.

This heatmap in \autoref{fig:hot} highlights two key insights. First, the distribution of values in $w$ is significantly sparse, indicating that many tokens in long-context input sequences are relatively unimportant. This reveals that evicting the KV pairs corresponding to these tokens can effectively reduce the KV cache. Second, $w$ varies significantly across layers with the complex hierarchical structure. We refer to this differences as Dynamic Differences of Importance Distribution (DDID). Since $w$ serves as the crucial guidance for KV eviction, the existence of DDID clearly shows the limitations of existing static KV cache eviction methods. They fail to capture this dynamic differences. Thus, to achieve more efficient and reasonable KV cache reduction, we should design a new eviction framework based on DDID.

\begin{figure}
    \centering
    \includegraphics[width=1\linewidth]{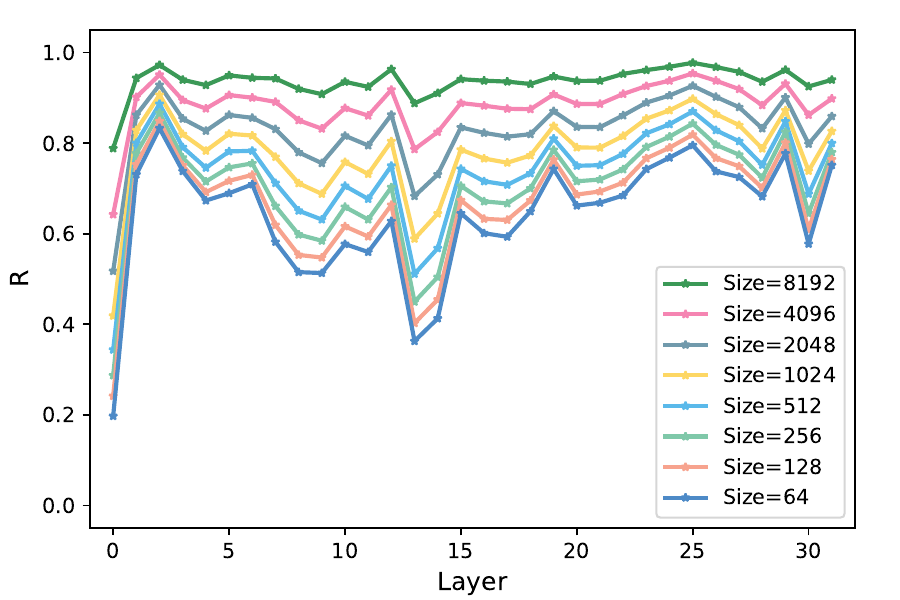}
    \caption{The variation of $R$ across layers under different KV cache sizes}
    \label{fig:fixsize}
\end{figure}

\begin{figure*}
    \centering
    \includegraphics[width=1\linewidth]{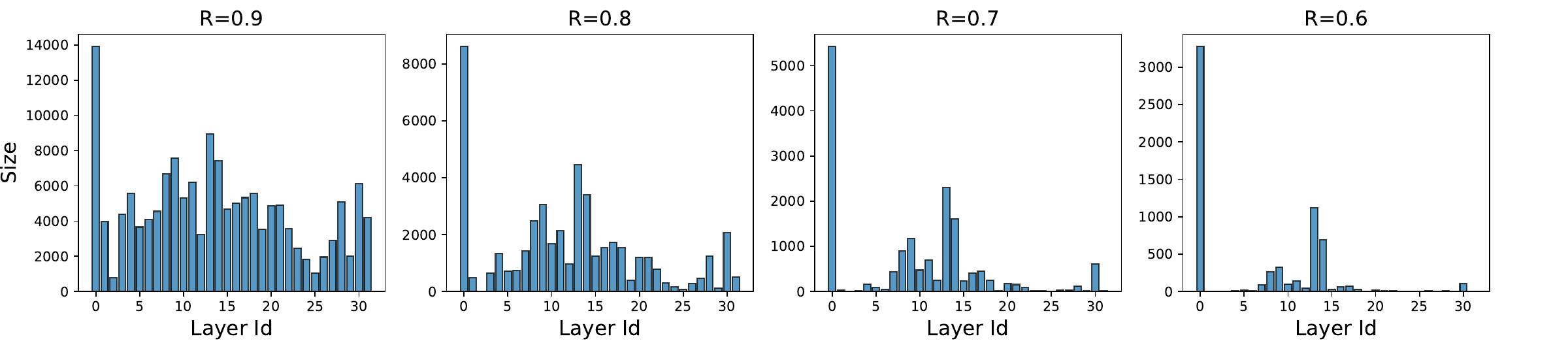}
    \caption{The variation of KV cache size across layers when the $R$ is fixed}
    \label{fig:fixR}
\end{figure*}

To further mathematically quantify the layer-wise differences reflected in DDID, we define a quantitative metric called Importance Retention Ratio ($R$). This metric $R$ represents the proportion of importance retained in a layer after evicting a specified number of tokens (with 100\% when no eviction). In particular, the formula for calculating $R_i$ of the $i$-th layer is given in Eq.~\eqref{Ri}.
\begin{equation}\label{Ri}
R_i=\frac{w^i_{retained}}{w^i_{original}}=\frac{Sum\left( Topk\left( n_{i,}w_i \right) \right)}{Sum\left( w_i \right)}\times 100\%
\end{equation}
Here, $w_i$ is the vector, indicating the distribution of attention scores at the $i$-th layer, and $n_i$ denotes the number of retained tokens, which corresponds to the size of the KV cache allocated to that layer. We define the ratio of $n$ to the input sequence length as the KV cache compression ratio. We pre-set fixed KV cache size $n$ which is a common approach adopted by most existing static methods, to observe how $R$ changes across layers. \autoref{fig:fixsize} reports some important results. $R$ across layers exhibits significant differences, which becomes increasingly large as the KV cache compression ratio decreases. By contrast, We pre-set a fixed $R$ to observe how $n$ accordingly changes across layers as shown in \autoref{fig:fixR}. We also observe the significant variation of the required cache size. Specifically, some layers, such as the 3rd and the 25th layers, achieve relatively ideal importance retention ratios even with a smaller KV cache size. However, in other layers, such as the first and the 13th layers, even with a large KV cache size allocated, the importance retention ratio still remains relatively small. We refer to this phenomenon as the sensitivity differences in importance retention ratio across layers w.r.t. the KV cache size, which is quantitative performance of DDID.

\subsection{Impact of DDID on Memory Costs and Inference Accuracy}
We conduct a theoretical analysis of the limitations of existing static methods in the presence of DDID and the potential of leveraging DDID for more efficient KV cache reduction. For the sake of analysis, we consider a simple single-head two-layer model. We abstract the process of performing KV eviction during the prefill stage as a special generation process in Eq.~\eqref{4} and Eq.~\eqref{5}.
\begin{gather}\label{4}
\scalebox{0.97}{$x_{t+1}^{1} = F\left( a_1 \right),$}\\
\scalebox{0.97}{$a_1=f\left( softmax \left( X_{\text{ows}}^{1} W_{Q}^{1} {W_{K}^{1}}^{T} {X_{t-\text{ows}}^{1}}^{T} \right) \right) {X_{t-\text{ows}}^{1}}^{T} W_{V}^{1} W_{O}^{1}$} \notag
\end{gather}
\begin{gather}\label{5}
\scalebox{0.97}{$x_{t+1}^{2} = F\left( a_2 \right),$} \\
\scalebox{0.97}{$a_2=f\left( softmax \left( X_{\text{ows}}^{2} W_{Q}^{2} {W_{K}^{2}}^{T} {X_{t-\text{ows}}^{2}}^{T} \right) \right) {X_{t-\text{ows}}^{2}}^{T} W_{V}^{2} W_{O}^{2}$} \notag
\end{gather}
$X_{ows}^{1}\in \mathbb{R}^{ows\times d}$ denotes the set of observation window tokens in the first layer, $ows$ is the size of the observation window and $d$ is the dimension of the token vector. Accordingly, $X_{t-ows}^{1}\in \mathbb{R}^{(t-ows)\times d}$ represents the set of tokens excluding the observation window tokens in the first layer, $t$ is the length of the input sequence. $W_{Q}^{1}$, $W_{K}^{1}$, $W_{V}^{1}$ $\in \mathbb{R}^{d\times p}$ and $W_{O}^{1}\in \mathbb{R}^{p\times d}$ are the weight in the first layer. Lastly, $f:\mathbb{R}^{ows\times p}\rightarrow \mathbb{R}^{1\times p}$ denotes a linear transformation for dimensionality reduction and $F:\mathbb{R}^{1\times d}\rightarrow \mathbb{R}^{1\times d}$ denotes the MLP block following attention block. The process abstracted by Eq.~\eqref{4} and Eq.~\eqref{5} is different from the actual generation process. The $X_{\text{ows}}^{2}$ in Eq.~\eqref{5} does not depend on the output of Eq.~\eqref{4}, which is a characteristic of the token selection process during the prefill stage. $W_{attn}=softmax \left( X_{\text{ows}} W_{Q} {W_{K}}^{T} {X_{t-\text{ows}}}^{T} \right)$ is the key metric. Eviction is to find $X\subset X_{t-\text{ows}}, X\in \mathbb{R}^{n\times d}, n$ is the size of the KV cache of this layer. It is not difficult to see that $W_{attn} \propto X$ and $x_{t+1} \propto W_{attn}$ for each layer. According to Eq.~\eqref{W} and Eq.~\eqref{Ri}, Eq.~\eqref{4} and Eq.~\eqref{5} can be written as follow:
\begin{equation}\label{6}
x_{t+1}^{1} = F\left( a_1 \right), \quad a_1 = w^1_{retained} {X_{1}}^{T} W_{V}^{1} W_{O}^{1} \propto  R_{1}
\end{equation}
\begin{equation}\label{7}
x_{t+1}^{2} = F\left( a_2 \right), \quad a_2 = w^2_{retained} {X_{2}}^{T} W_{V}^{2} W_{O}^{2} \propto R_{2}
\end{equation}
To better describe the distribution differences in \autoref{fig:hot}, we define Importance to Size Ratio ($ISR$) to describe the relationship between each layer's $R$ and the KV cache size $n$. The definition of $ISR$ at the $i$-th layer is as follows in Eq.~\eqref{8}.
\begin{equation}\label{8}
ISR_i=\frac{R_i}{\log _2n_i}\,\,\left( n_i>1 \right) 
\end{equation}
$ISR_i$ is directly related to $n_i$. We calculate the difference of $ISR$ for two layers. The formula for calculating the difference $f(n_i)$ is as follows in Eq.~\eqref{9}.
\begin{equation}\label{9}
\begin{split}
     f\left( n_i \right) &=\frac{ISR\left( n_i \right) -ISR\left( n_i-1 \right)}{\left( n_i \right) -(n_i-1)}\\
     &=\frac{R_i\left( n_i \right)}{\log _2\left( n_i \right)}-\frac{R_i\left( n_i-1 \right)}{\log _2\left( n_i-1 \right)} 
\end{split}
\end{equation}
Let $f\left( n_i \right)=\delta$, where $\delta$ is the defined threshold. When the difference reaches the threshold, it is considered that the boundary between red and white in \autoref{fig:hot} has been reached at this time. Now, DDID between two layers can be described as: let $f_1\left( n_1 \right)=\delta$ and $f_2\left( n_2 \right)=\delta$ , where $n_1>n_2$ or $n_1<n_2$. Here, we assume it is the former. From the perspective of reducing KV cache size, our key goal is to reduce $n_2$. Consider the situation where $n_1=n_2$, $f_1\left( n_1 \right)=\delta$, $f_2\left( n_2 \right)<\delta$, as observed in existing methods. Let $f_2\left( n \right)=\delta$, where $n=n_{2th}$. $n_{2th}$ is threshold in the second layer. Then, we carry out the following transformations on $f\left( n_2 \right)$:
\begin{equation}
    f\left( n_2 \right) =\frac{R_2\left( n_2 \right)}{\log _2\left( n_2 \right)}-\frac{R_2\left( n_2-1 \right)}{\log _2\left( n_2-1 \right)}=\delta 
\end{equation}
\begin{equation}
R_2\left( n_2-1 \right) =\frac{\log _2\left( n_2-1 \right) R_2\left( n_2 \right)}{\log _2\left( n_2 \right)}-\delta \log _2\left( n_2-1 \right)
\end{equation}
\begin{equation}
R_2\left( n_2-1 \right) \approx R_2\left( n_2 \right) -\delta \log _2\left( n_2-1 \right)
\end{equation}
Let $\varDelta=\delta \log _2\left( n_2-1 \right) $, where $\varDelta$ is a very small value. The approximation is as follows:
\begin{equation}
    R_2\left( n_2-1 \right) \approx R_2\left( n_2 \right) -\varDelta \approx R_2\left( n_2 \right)
\end{equation}
According to Eq.~\eqref{Ri} and Eq.~\eqref{7}, $w^2_{retained}(n_2-1) \approx w^2_{retained}(n_2)$. For the second layer, within the acceptable loss range $\varDelta$, at least one set of such recursive relationships can surely be found between $n_2$ and $n_{2th}$. This is the potential direction for reducing the KV cache size. From the perspective of improving inference accuracy, we allocate the space saved in the second layer to the first layer. Assume the situation where $n_1=n_2$, $f_1\left( n_1+1 \right)>\delta$, $f_2\left( n_2 \right)=\delta$. Subsequently, we perform the following transformations on $f_1\left( n_1+1 \right)>\delta$:
\begin{equation}
    f\left( n_1+1 \right) =\frac{R_1\left( n_1+1 \right)}{\log _2\left( n_1+1 \right)}-\frac{R_1\left( n_1 \right)}{\log _2\left( n_1 \right)}>\delta 
\end{equation}
\begin{equation}
    R_1\left( n_1+1 \right) - R_1\left( n_1 \right) > \delta \log _2\left( n_1+1 \right)
\end{equation}
In the above analysis of memory optimization, we can get $R_2\left( n_2 \right) - R_2\left( n_2-1 \right) < \delta \log _2\left( n_2-1 \right)$. Because $a_1+a_2\propto R_1+R_2$, $R_1\left( n_1+1 \right) - R_1\left( n_1 \right)+R_2\left( n_2-1 \right) - R_2\left( n_2 \right)>0$, which indicates that there must exist an allocation $\{n_1+1,n_2-1\}$ that is superior to the original allocation$\{n_1,n_2\}$ for prefill phase. Combined with the proven hypothesis of attention persistence\cite{NEURIPS2023_aScissorhands}, KV of these retained tokens are also important for subsequent inferences, thereby enhancing the final inference effect. Therefore, leveraging DDID for personalized KV cache reduction is crucial.

\begin{figure}
    \centering
    \includegraphics[width=1\linewidth]{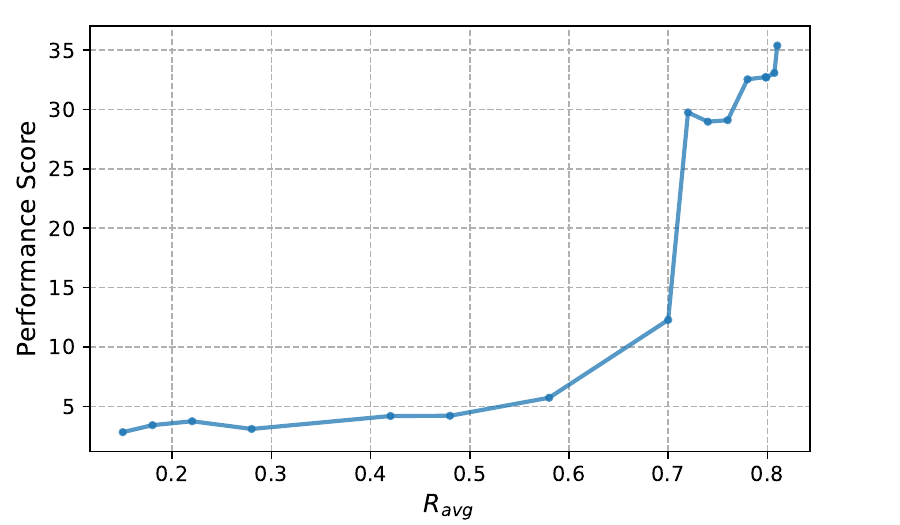}
    \caption{The relationship between $R_{avg}$ and inference results}
    \label{fig:Ravg}
\end{figure}

To better explore personalized KV cache reduction strategy, we further investigate the the impact of DDID on inference results. In theory, the larger the value of $R$ for each layer, the more beneficial it is for inference results. Due to the relative independence of input between layers in the token selection process demonstrated by Eq.~\eqref{4} and Eq.~\eqref{5}, we hypothesize that the larger the total sum of $R$ across all layers, the more favorable it is for the final end-to-end inference results. 
$R_{avg}$ serves as a good reflection of the overall sum of $R$ across all layer structures. We have experimentally validated this relationship between $R_{avg}$ and inference performance results, as shown in \autoref{fig:Ravg}. Under a constant total KV cache size across all layers, we manually adjust the allocation of KV cache size for each layer. As the allocation scheme varies, the value of 
$R$ for each layer changes, which in turn affects the $R_{avg}$. The relationship between $R_{avg}$ and inference results inspires us to design a novel KV cache reduction strategy.

\begin{figure*}
    \centering
    \includegraphics[width=0.9\linewidth]{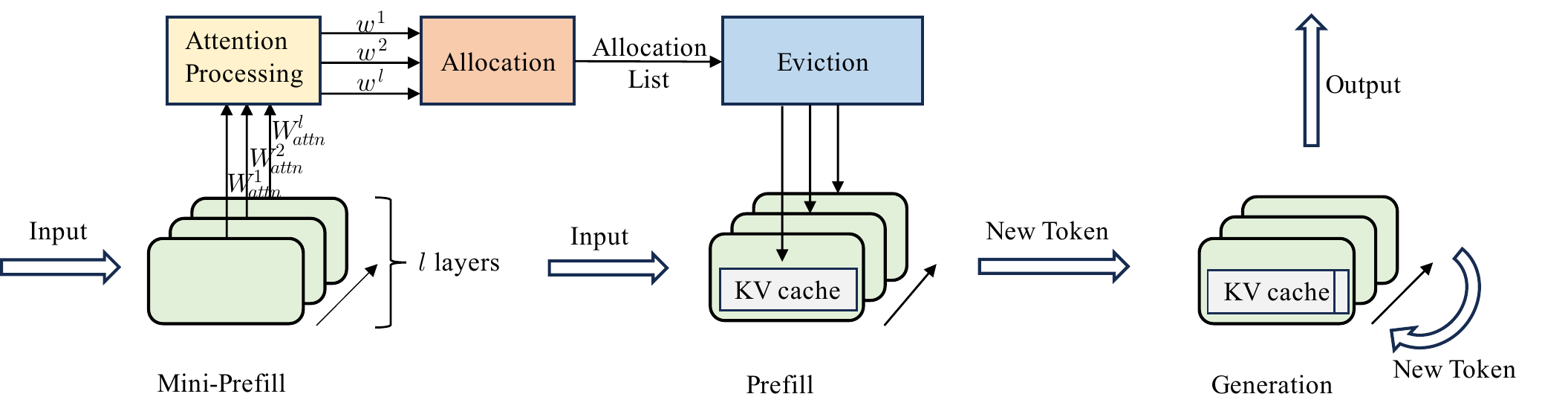}
    \caption{The inference procedures under XKV}
    \label{fig:framework}
\end{figure*}

\subsection{DDID-driven Memory Reduction}
For some layers where $R_i$ is highly sensitive to $n_i$, even with a small $n_i$, we can still obtain a considerably large $R_i$, leading to a significant increase in $R_{avg}$. For such layers, it is worthwhile to allocate more memory resources, i.e., a large cache size. This is because a small memory consumption can result in a substantial improvement in $R_{avg}$. Conversely, for layers where $R_i$ is less sensitive to $n_i$, even though we allocate significantly more cache resources than other layers, the improvement in $R_i$ is still limited, let alone any substantial impact on $R_{avg}$. For these layers, the memory cost is not proportional to the benefit, so it would be better to reduce the KV cache size w.r.t. them. We can allocate the limited resources for other layers to increase their KV cache size, since these layers can contribute significantly to $R_{avg}$. This is a straightforward heuristic approach. However, it is difficult to exactly describe this strateg, making it challenging to quantitatively determine the globally optimal solution. Inspired by the DDID research, we have a direction for developing a strategy that adaptively reduces KV cache. We model the determination of KV cache reduction strategy as an optimization problem under two constraints:
\begin{equation}\label{opproblem1}
    \begin{split}
&argmax \left( \sum_{i=1}^l{\frac{Sum\left( Topk\left( n_i,w_i \right) \right)}{Sum\left( w_i \right) l}} \right)\\
&s.t. \quad \quad \quad \sum_{i=1}^l{n_i}=Total\,\,Size
    \end{split}
\end{equation}
\begin{equation}\label{opproblem2}
    \begin{split}
&argmin \left(\sum_{i=1}^l{n_i}\right)\\
&s.t.R_{avg}=Set\,\,Value\\
    \end{split}
\end{equation}
By Eq.~\eqref{opproblem1} and Eq.~\eqref{opproblem2}, we know they are combinatorial optimization problems. Due to the finite solution space, we can definitely find a globally optimal solution. In particular, given the total memory resource budget for all layers, or the required accuracy-related value of $R_{avg}$, our goal is to find an optimal setting about the layer-specific KV cache size that maximizes $R_{avg}$, or minimizes the total memory consumption. This optimal setting represents the most prominent memory allocation for KV caches specific to every layer.

\section{XKV: A Lightweight KV-Cache Memory Reduction Framework}\label{sec:impl}
Based on Sec.~\ref{research}, we propose a novel framework XKV, which leverages DDID in hierarchical structure to enable personalized KV cache reduction of each layer. This approach aims to achieve more efficient memory overhead optimization. In this section, we introduce our approach by elaborating on three key parts: the framework design, the algorithmic implementation, and the strategy for reducing computational overhead.

\subsection{XKV Framework with Mini-Prefill}
By Eq.~\eqref{opproblem1} and Eq.~\eqref{opproblem2}, finding the optimal eviction strategy during inference is challenging. This is because at the $i$-th layer, the state of the $(i+1)$-th layer cannot be predicted, and the locally optimal solution obtained at the current layer may not necessarily lead to a globally optimal solution which is also the limitation of traditional heuristic algorithms. To address this issue, we propose XKV inference framework based on mini-prefill, as illustrated in \autoref{fig:framework}.

We divide the inference procedures into three phases: the mini-prefill phase, the prefill phase, and the generation phase. Mini-prefill is conducted before the prefill phase. During mini-prefill, inference computations are performed layer by layer based on the original input. In this process, the intermediate results $W_{attn}$ generated by each layer are processed by the attention processing module into corresponding attention score distribution vectors $w$. At the end of mini-prefill, these attention score distribution vectors for all layers are used to derive a globally optimal KV cache size allocation list through an adaptive KV cache size allocation algorithm. In the prefill phase, when caching KV for each layer based on the original input, the KV cache is dynamically compressed according to the KV cache size allocation list generated during mini-prefill. By the end of the prefill phase, the GPU memory holds the reduced KV cache. Finally, during the generation phase, starting from the first token produced in the prefill phase, new tokens are generated autoregressively based on the KV cache. The KV corresponding to the newly generated tokens are also stored in the KV cache, and this process continues until inference is complete.
\begin{figure}
    \centering
    \includegraphics[width=0.825\linewidth]{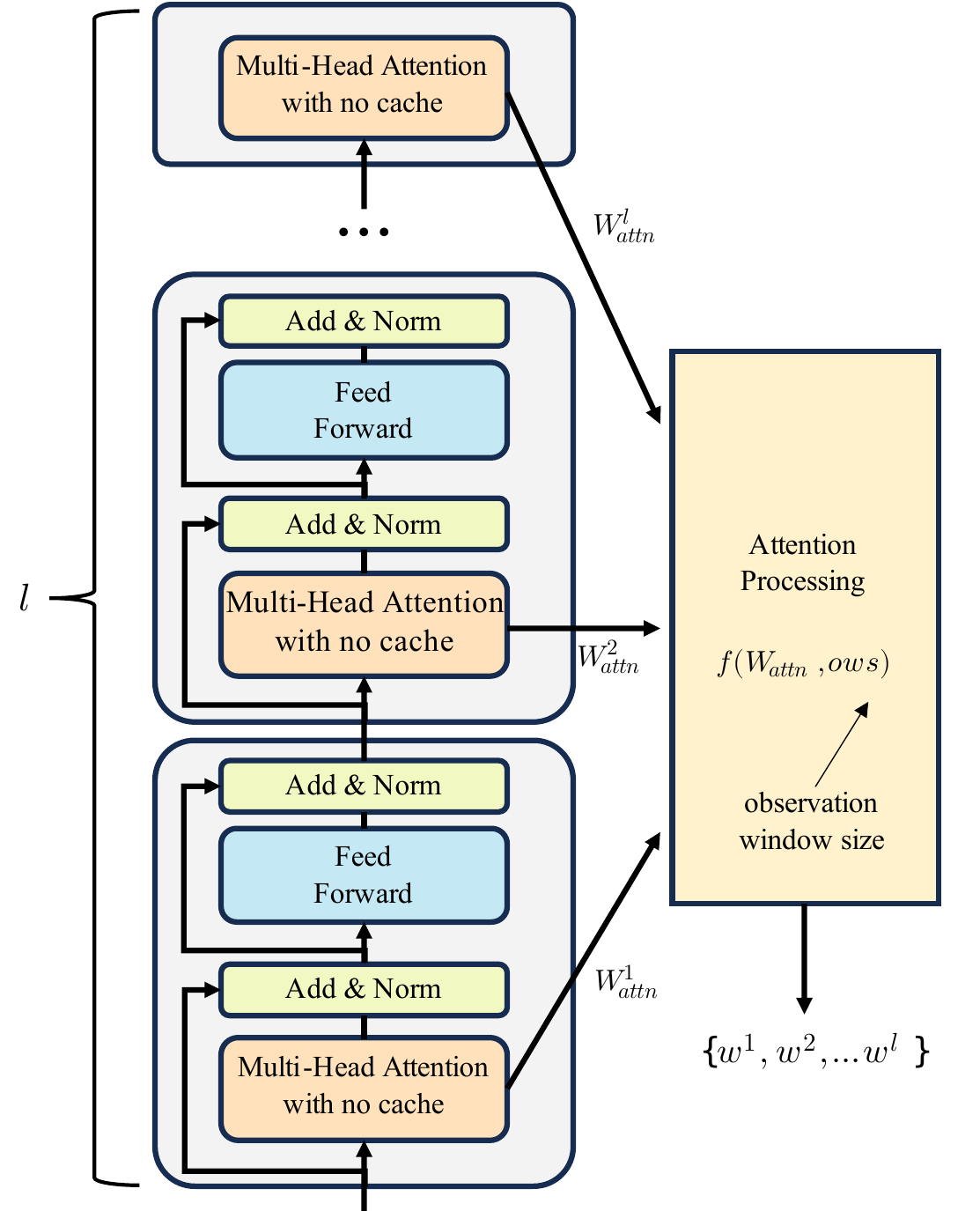}
    \caption{The architecture of mini-prefill}
    \label{fig:miniprefill}
\end{figure}
\begin{figure}
    \centering
    \includegraphics[width=0.825\linewidth]{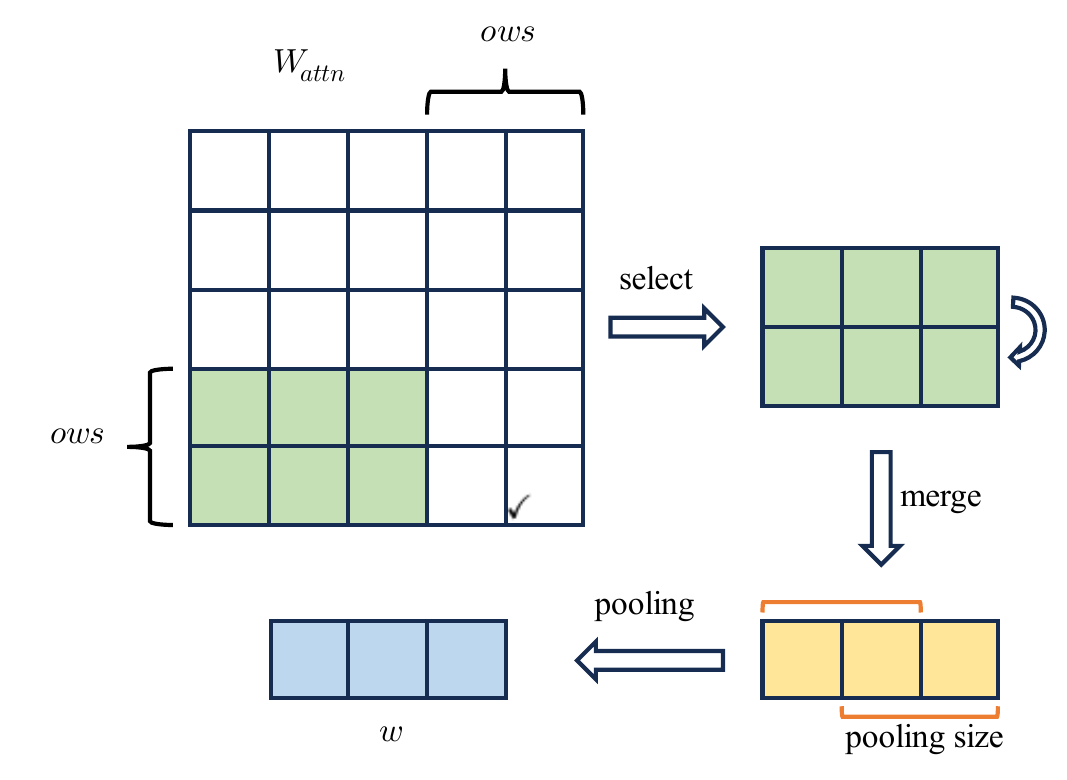}
    \caption{The attention processing procedure}
    \label{fig:attnpro}
\end{figure}

Mini-prefill is a computation-intensive yet lightweight module, with its architecture illustrated in \autoref{fig:miniprefill}. It consists of $l$ layers of Transformer blocks and an attention processing module. Each Transformer block comprises four primary components: a multi-head attention module, a feedforward neural network module, residual connections and layer normalization. The attention weight matrices required for our algorithms are intermediate results produced by the multi-head attention module in each Transformer layer. To obtain these matrices, they are temporarily saved during inference as each layer's multi-head attention module is executed. These matrices are then processed by the attention processing module using the specialized operations described in Eq.~\eqref{W}, resulting in attention score distribution vectors $w$ for each layer. The detailed attention processing flow is depicted in \autoref{fig:attnpro}.The original matrix $W_{attn}$ is a $t\times t$ square matrix, where $t$ is the length of original input sequence. First, it is selected based on the observation window size $ows$, resulting in a $ows\times (t-ows)$ matrix. This matrix is then averaged along its first dimension, producing a $1\times (t-ows)$ vector. Finally, pooling is applied to this vector based on the pooling size, yielding the attention score distribution vector.

It is worth noting that, to minimize unnecessary memory and computational overhead in mini-prefill, no KV caching is performed in the multi-head attention modules, unlike in the prefill phase. The first $l-1$ Transformer layers of mini-prefill consist of the standard four components, but the $l$-th layer includes only the multi-head attention module. This design means that when the final multi-head attention module of the $l$th layer is completed, the entire mini-prefill process terminates immediately. Subsequent operations, such as residual connections, layer normalization, and the feed forward neural network module, are skipped because all the required information has already been obtained at this time.

\subsection{Personalized \& Fast Memory Allocation}
\begin{figure}[h]
    \centering
    \includegraphics[width=0.8\linewidth]{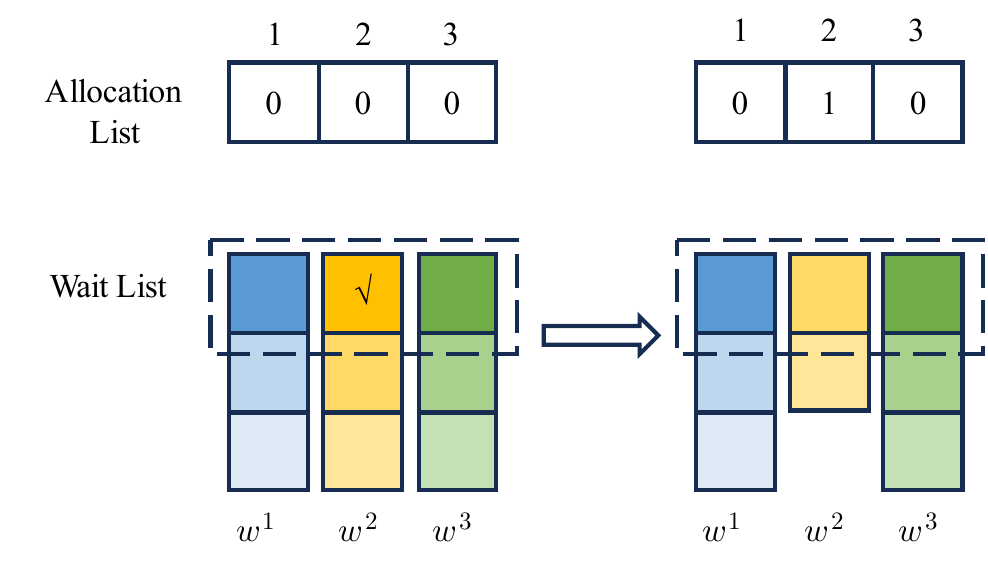}
    \caption{Illustration of procedures in a single iteration}
    \label{fig:algorithm}
\end{figure}

To fully exploit DDID in the attention score distribution vectors $w$, we aim to solve the combinatorial optimization problem described in Eq.~\eqref{opproblem1} and Eq.~\eqref{opproblem2} to obtain its globally optimal solution. This globally optimal solution can be regarded as the optimal KV cache reduction strategy derived based on the DDID. However, solving such a combinatorial optimization problem is often challenging due to its NP-hard nature, making direct computation infeasible in most cases. Fortunately, leveraging the unique discrete properties of the objective function, we have developed an efficient algorithm for adaptive KV cache size allocation as described in Algorithm \ref{suanfa}. This algorithm effectively computes the globally optimal KV cache reduction strategy. The proposed algorithm adopts a greedy strategy to incrementally identify the layer and corresponding token that contribute the most to $R_{avg}$. Although the iterative objectives vary slightly under different constraints, the ultimate goal remains the same: to obtain an optimal allocation list. The core idea of the algorithm lies in maintaining and dynamically updating two critical lists: the allocation list and the wait list. The allocation list records the current KV cache size for each layer, while the wait list tracks the token within each layer that presently offers the largest contribution to improving $R_{avg}$. By iteratively updating these two lists, the algorithm incrementally optimizes the KV cache allocation to achieve an optimal solution. \autoref{fig:algorithm} illustrates the procedures of a single iteration. We take the constraint conditions in Eq.~\eqref{opproblem1} as an example. The detailed process of the algorithm is as follows. Initially, two lists are prepared: the allocation list and the wait list. The allocation list is initialized to zero for all layers, indicating that no KV cache has been allocated to any layer at the beginning. Meanwhile, the wait list is initialized with the token in each layer that makes the largest contribution to $R_{avg}$. After initialization, the algorithm proceeds to an iterative optimization phase, which consists of $N$ iterations. In each iteration, the algorithm identifies the element with the largest contribution value from the wait list. Based on this maximum value, it determines the corresponding layer index. Then, the KV cache size of the identified layer in the allocation list is incremented by 1. Subsequently, using the updated KV cache size, the algorithm reevaluates the token set in the same layer to select the next token that makes the largest contribution to 
$R_{avg}$. The wait list is then updated with this new token value. Through this iterative process, the algorithm dynamically adjusts the allocation list, progressively approaching the optimal solution.
\begin{center}
	\begin{minipage}{1\linewidth} 
        \begin{algorithm}[H]
            \caption{Adaptive Allocation Algorithm}\label{suanfa}
              \begin{algorithmic}[1]
                \REQUIRE Total Size $N$ or Set Value $R_{avg}$,  \\\quad \, Number of Layers $L$,\,$W = \{w_1, w_2, \ldots, w_l\}$
                \ENSURE Allocation\,List $List_A= \{n_1, n_2, \ldots, n_l\}$
                \STATE \text{initialize} Allocation\,List $List_A=\left\{ 0,0,...,0 \right\}$ 
                \STATE \text{initialize} Wait List $List_W=\left\{ 0,0,...,0 \right\}$
                                \FOR{$i$ in range(L)}
                    \STATE  $List_W[i]\gets$ Sum$(w_i[-1:])/$sum$(w_i)$
                \ENDFOR
                \STATE Sort($W$)
                \IF{\text{Exist($N$)}}
                \FOR{$i$ in range(N)}
                    \STATE  maximum $\gets$ Max($List_W$)
                    \STATE  index $\gets$ Getindex($List_W$, maximum)
                    \STATE  $List_A[$index$]$ $\gets$ $List_A[$index$]$$+1$
                    \STATE  pos $\gets$ $List_A[$index$]$
                    \STATE  $List_W[$index$]$$\gets$ Sum$(w_i$[$-($pos$+1):$pos$])/$Sum$(w_i)$
                \ENDFOR
                \ENDIF
                \IF{\text{Exist($R_{avg}$)}}
                \STATE \text{initialize} $R_{total}$ = 0
                \FOR{$R_{total}/L<R_{avg}$}
                    \STATE  maximum $\gets$ Max($List_W$)
                    \STATE  $R_{total}$ $\gets$ $R_{total}$ $+$ maximum
                    \STATE  index $\gets$ Getindex($List_W$, maximum)
                    \STATE  $List_A[$index$]$ $\gets$ $List_A[$index$]$$+1$
                    \STATE  pos $\gets$ $List_A[$index$]$
                    \STATE  $List_W[$index$]$$\gets$ Sum$(w_i$[$-($pos$+1):$pos$])/$Sum$(w_i)$
                \ENDFOR
                \ENDIF
                \RETURN Allocation\,List $List_A$
            \end{algorithmic}
		\end{algorithm}
	\end{minipage}
\end{center}
Proof: Considering a two-layer model, the total KVcache size for the two layers is set to $N$, $n_1+n_2=N$. To prove the optimality of Algorithm \ref{suanfa}. Suppose there is an optimal solution other than Algorithm \ref{suanfa}, Allocation List1 = \{$n_1,n_2$\}. $R_{avg1}$ is calculated as Eq.~\eqref{19}.
\begin{equation}\label{19}
\frac{Sum\left( Topk\left( n_1,w_1 \right) \right)}{2Sum\left( w_1 \right)}+\frac{Sum\left( Topk\left( n_2,w_2 \right) \right)}{2Sum\left( w_2 \right)}
\end{equation}
Since Allocation List1 is not obtained according to the strategy of Algorithm \ref{suanfa}, it means that the top $N$ tokens in the two layers are not completely selected. According to the greedy decision of Algorithm \ref{suanfa}, we must be able to obtain a feasible solution Allocation List2 on the basis of Allocation List1. Allocation List2=$\{n_1+1,n_2-1\}$ or \{$n_1-1,n_2+1$\}. Here we assume that Allocation List2=\{$n_1+1,n_2-1$\}. $R_{avg2}$ is calculated as Eq.~\eqref{20}.
\begin{equation}\label{20}
    \frac{Sum\left( Topk\left( n_1+1,w_1 \right) \right)}{2Sum\left( w_1 \right)}+\frac{Sum\left( Topk\left( n_2-1,w_2 \right) \right)}{2Sum\left( w_2 \right)}
\end{equation}
The calculation of $R_{avg2}-R_{avg1}$ is shown in Eq.~\eqref{21}:
\begin{equation}\label{21}
    \begin{split}
& \frac{Sum\left( Topk\left( n_1+1,w_1 \right) \right) -Sum\left( Topk\left( n_1,w_1 \right) \right)}{2Sum\left( w_1 \right)}+  \\ & \frac{Sum\left( Topk\left( n_2-1,w_2 \right) \right) -Sum\left( Topk\left( n_2,w_2 \right) \right)}{2Sum\left( w_2 \right)} 
\\
=& \frac{Sort\left( w_1 \right) \left[ -\left( n_1+1 \right) \right]}{2Sum\left( w_1 \right)}-\frac{Sort\left( w_2 \right) \left[ -n_2 \right]}{2Sum\left( w_2 \right)}
    \end{split}
\end{equation}
Since Allocation List2=$\{n_1+1,n_2-1\}$ is obtained according to Algorithm \ref{suanfa}, it must exist $\frac{Sort\left( w_1 \right) \left[ -\left( n_1+1 \right) \right]}{Sum\left( w_1 \right)}$ $>$ $\frac{Sort\left( w_2 \right) \left[ -n_2 \right]}{Sum\left( w_2 \right)}$. Consequently $R_{avg2}-R_{avg1}$ $>$ 0. This indicates that we can find a solution Allocation List2, which is no worse than the original solution Allocation List1 based on Algorithm \ref{suanfa}. This proves the optimal substructure property, thus concluding the proof of the optimality of Algorithm \ref{suanfa}.
\begin{center}
        \begin{algorithm}[H]
            \caption{Eviction Algorithm of the $i$-th Layer}\label{suanfa2}
              \begin{algorithmic}[1]
                \REQUIRE KV cache Size $n_i$, Observation Window Size $ows$,\\ \, \, \, \,Pooling Size $ps$, $Q_i$, $K_i$, $V_i$
                \ENSURE $K_i$, $V_i$
                \STATE  $W_{attn}^i$ $\gets$ $softmax((Q_iK_i^T)/\sqrt{d_k})$
                \STATE  $w_i$ $\gets$ Select$(W_{attn}^i, ows)$
                \STATE  $w_i$ $\gets$ Merge$(w_i, ows)$
                \STATE  $w_i$ $\gets$ Pooling$(w_i, ps)$
                \STATE  index $\gets$ TopK$(w_i, n_i)$
                \STATE  $K_i$ $\gets$ EvictByIndex$(K_i, index)$
                \STATE  $V_i$ $\gets$ EvictByIndex$(V_i, index)$
                \RETURN $K_i$, $V_i$
            \end{algorithmic}
		\end{algorithm}
\end{center}

During the prefill phase, each layer executes the multi-head attention module and applies a dynamic eviction algorithm for KV cache reduction, guided by the allocation list obtained from Algorithm \ref{suanfa}. The eviction process for the $i$th layer is detailed in Algorithm \ref{suanfa2}. Specifically, Algorithm \ref{suanfa2} computes the attention weight through $Q$ and $K$ and performs a specialized attention operation, as illustrated in \autoref{fig:attnpro}. Based on the KV cache size $n_i$ allocated to the $i$-th layer, a top-k algorithm is employed to identify the indices of the top $n_i$ most important tokens for this layer. These $n_i$ tokens are deemed to have a persistent impact on inference and represent the globally optimal choice for retention at this layer. Subsequently, the KV pairs corresponding to these selected token are retained in the GPU memory, while the rest are evicted. During the subsequent generation phase, the attention mechanism leverages the cached KV pairs for efficient computation.

\subsection{Sampling-based Overhead Reduction}
\begin{figure}
    \centering
    \includegraphics[width=1\linewidth]{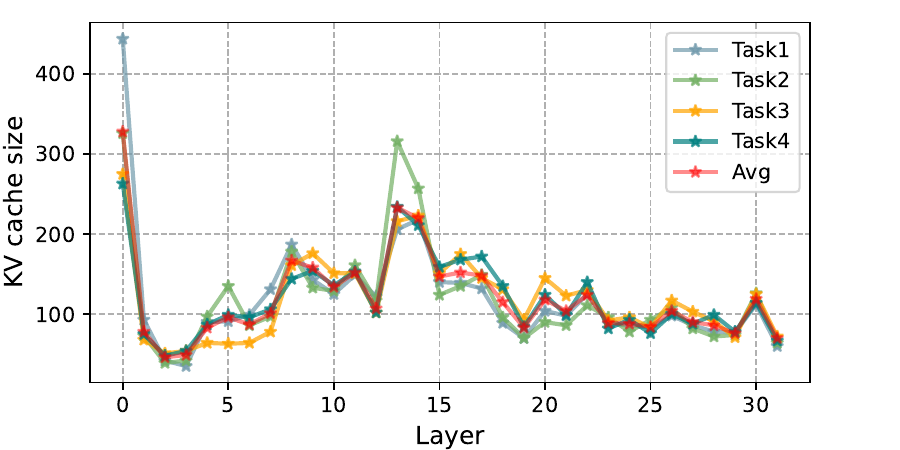}
    \caption{The KV cache size of each layer in QA tasks}
    \label{fig:43}
\end{figure}
We consider two distinct inference scenarios: offline scenarios and online scenarios. In offline scenarios, where the latency requirements for individual inference tasks are relatively relaxed, our proposed method can be applied to each task. It adaptively allocates KV cache sizes and performs dynamic KV cache eviction for each layer. In this context, a certain degree of additional computational overhead is acceptable, allowing for personalized optimization to achieve the best inference results. In contrast, for online scenarios, there are stricter latency constraints while maintaining acceptable inference performance, running mini-prefill and our algorithms is not
free in XKV. Fortunately we draw inspiration from the observations in \autoref{fig:43} and adopt a sampling strategy to significantly reduce computational overhead while still achieving satisfactory inference results. 

As illustrated in \autoref{fig:43}, we observe that for inference tasks of the same type, the optimal allocation of KV cache across layers exhibits a certain level of consistency: the variation in KV cache sizes follows similar cross-layer trends, although specific KV cache sizes for the same layer may vary slightly between tasks. Based on this observation, in scenarios with strict latency requirements, we can sample a subset of inference tasks from the same category. Using our proposed method, we can fully process these sampled tasks. During this phase, we record the KV cache allocation for each layer. After the sampling phase, we compute the average of the recorded allocation results. For subsequent inference tasks of the same type, we directly apply this averaged KV cache size allocation list for eviction, thereby balancing performance and efficiency.

\section{Evaluation}\label{sec:exp}
We finally evaluate the effectiveness of our XKV solution.

\subsection{Experimental Setting}
\paragraph{Datasets}LongBench\cite{bai-etal-2024-longbench} is a benchmark for comprehensive evaluation of LLMs' long-context understanding ability in a bilingual and multi-task setting. It covers key long-text applications, including single-document QA, multi-document QA, summarization, few-shot learning, and code completion. The inference accuracy metric is specific to the given application, including F1, Rouge-L, Accuracy, and Edit Similarity\footnote{For more details, please refer to https://github.com/THUDM/LongBench.}. For all metrics, the larger, the better. We select 14 datasets involved in the applications mentioned above.
\paragraph{Benchmark Model and Competitors} We run the advanced open-source LLama3-8b-instruct\cite{dubey2024llama3} as the representative LLM model. It has 8 billions parameters. We compare our XKV against 6 up-to-date competitors, including FullKV\cite{dubey2024llama3}, StreamingLLM\cite{xiao2024streamingllm}, H2O\cite{zhang2024h2o}, SnapKV\cite{li2024snapkv}, PyramidInfer\cite{yang-etal-2024-pyramidinfer}, and PyramidKV\cite{zhang2024pyramidkv}. FullKV provides the standard accuracy since no data is evicted. Table~\ref{t1} summarizes the features of other competitors. They evict data based on attention score. Among them, PyramidInfer, PyramidKV and our XKV can customize the cache resource allocation across layers. The former two employ the static settings for any application; while XKV can dynamically tune settings by our DDID insight.

\begin{table}
 \caption{Features of competitors in our experiments}\label{t2}
  \centering
 \resizebox{0.48\textwidth}{!}{
\begin{tabular}{ccccc}
\toprule
\makecell[l]{Methods} &
\makecell[c]{Attention \\score-based} &
\makecell[c]{Cross-layer\\allocation}& 
\makecell[c]{Utilization\\ of DDID} &
\makecell[c]{Dynamic \\tuning}  \\

\midrule
\makecell[l]{StreamingLLM, \\H2O, SnapKV}&$\checkmark$ &$\times$ &$\times$ &$\times$  \\
\midrule
\makecell[l]{PyramidInfer, \\PramidKV}&$\checkmark$ &$\checkmark$ &$\times$ &$\times$  \\
\midrule
\makecell[l]{XKV(Ours)}&$\checkmark$ &$\checkmark$ &$\checkmark$ &$\checkmark$  \\
\bottomrule
\end{tabular}
  }
\end{table}

\paragraph{Other Settings} By default, we set the observation window size, the sampling ratio, and the pooling size respectively as 8, 10\% and 7. We particularly use average computation as the pooling method. Following PyramidInfer\cite{yang-etal-2024-pyramidinfer}, we also use the compression ratio to indicate the effectiveness of memory reduction. It is the ratio of the total cache size of all layers after eviction, to the original one. A low ratio indicates that only a few K-V pairs are preserved, i.e., a significant memory reduction. Finally, all experiments are conducted on a NVIDIA RTX A6000 GPU (48GB) device. For each kind of inference task, we run all associated datasets and report the average evaluation metric.

Below we explore the performance comparison in terms of memory reduction (Sec.~\ref{sec:exp:memory}), inference accuracy (Sec.~\ref{sec:exp:accuracy}), and inference efficiency (Sec.~\ref{sec:exp:runtime}).

\begin{figure*}
    \centering
    \includegraphics[width=0.9\linewidth]{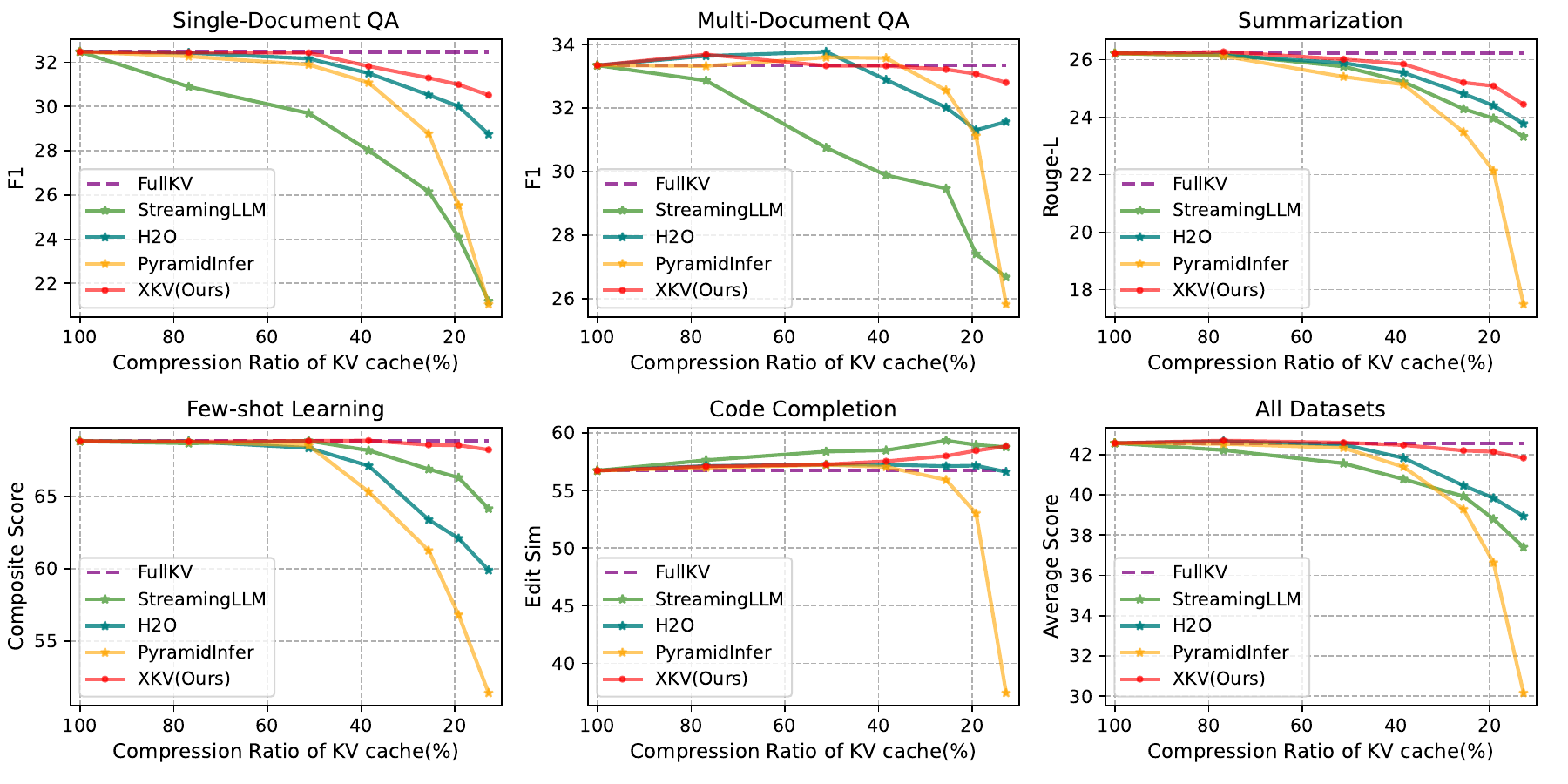}
    \caption{Comparison on memory reduction under different accuracy loss bounds}
    \label{fig:51}
\end{figure*}

\subsection{Evaluation on Memory Reduction}\label{sec:exp:memory}

\begin{figure*}
    \centering
    \includegraphics[width=0.9\linewidth]{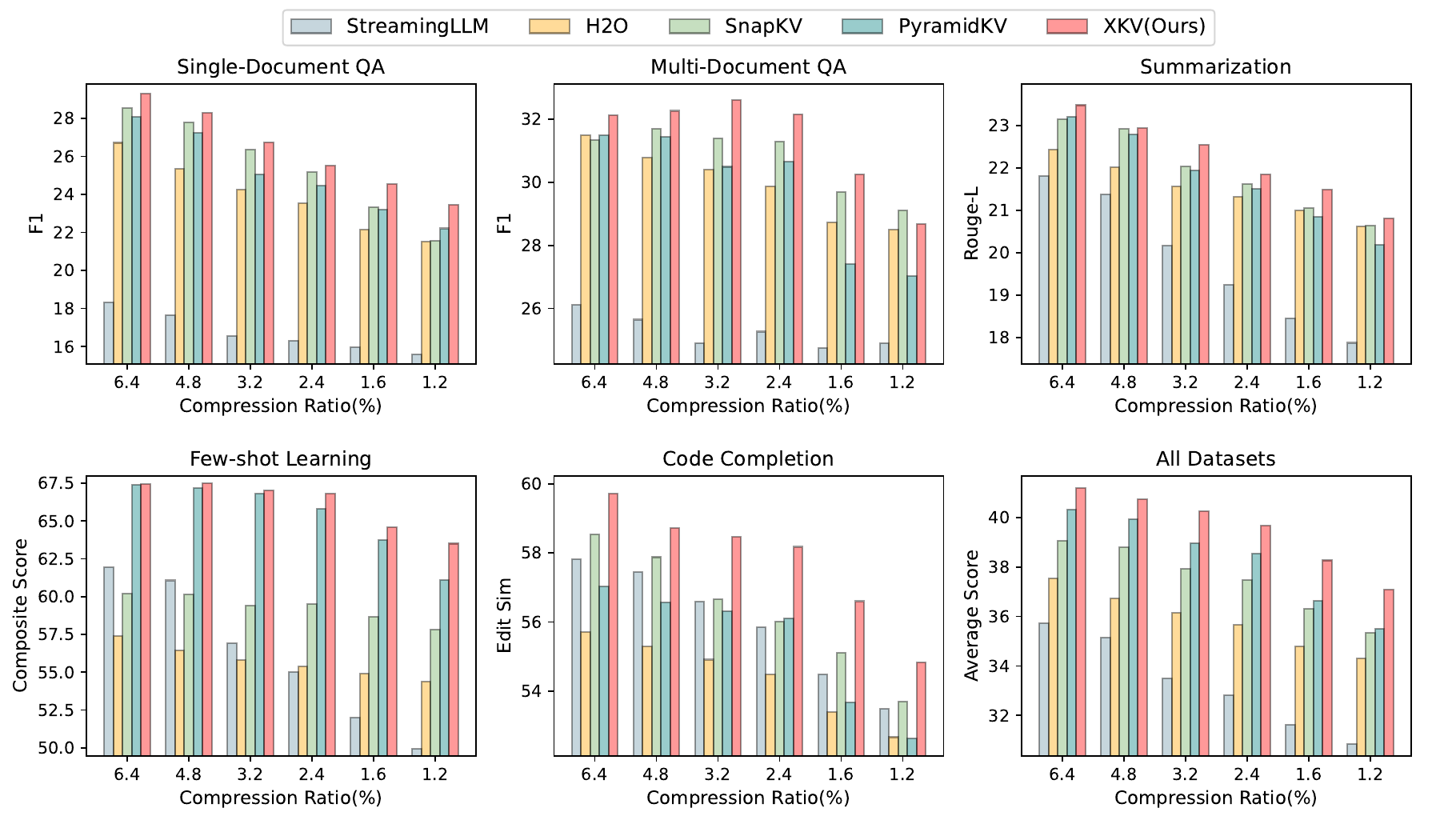}
    \caption{Comparison on inference accuracy under different memory constraints}
    \label{fig:52}
\end{figure*}

\begin{table*}
 \caption{The specific inference performance of each method on 14 datasets}
  \centering
  \resizebox{\textwidth}{!}{
  \begin{tabular}{llllllllllllllll}
    \toprule
    \multirow{2}{*}{Methods} &   \multicolumn{3}{c}{Single-Document QA} & \multicolumn{3}{c}{Multi-Document QA} & \multicolumn{3}{c}{Summarization} & \multicolumn{3}{c}{Few-shot Learning}   & \multicolumn{2}{c}{Code} &\multirow{2}{*}{Avg} \\ 
    \cmidrule(r){2-4} 
    \cmidrule(r){5-7} 
    \cmidrule(r){8-10}
    \cmidrule(r){11-13}
    \cmidrule(r){14-15} 
    &NaQA   &QAs    &MFQA   &HPQA   &WKQA   &Mus    &GovR   &Qmsum  &Mnews  &Trec   &TrivQA &Ssum   &Lcc    &RPB\\
    \midrule
    \multicolumn{16}{c}{KV cache Compression Ratio = 1.6\%}\\
    \midrule
StreamingLLM&
18.13 & 8.51 & 21.31 & 32.86 & 25.85 & 15.51 & 16.76 & 20.43 & 18.16 & 45.00 & 74.59 & 36.35 & 55.98 & 53.00&31.60\\
H2O
&21.92& 15.02& 29.45& 39.59& 27.30& 19.32& 20.30& 22.57& 20.12& 35.50& 89.65& \textbf{39.53}& 55.67& 51.10& 34.79\\
SnapKV
&22.59& 16.11& 31.23& 40.80& 29.01& 19.24& 19.82& 21.98& 21.35& 48.00& 89.32& 38.68& 57.43& \textbf{52.77}& 36.31\\
PyramidKV
&21.80&\textbf{16.82}& 30.97& 37.80& 25.93& 18.53& 19.95& 21.90& 20.67& 65.00& 87.83& 38.42& 56.03& 51.32& 36.64\\
XKV(Ours)
&\textbf{22.73}& 16.32& \textbf{34.53}& \textbf{41.59}& \textbf{29.28}& \textbf{19.91}& \textbf{20.45}& 22.44& \textbf{21.59}& \textbf{65.00}& \textbf{89.90}& 38.80& \textbf{58.52}& \textbf{54.68}& \textbf{38.27}\\
    \bottomrule
  \end{tabular}
  }
  \label{tab:table3}
\end{table*}

\begin{table*}
  \centering
  \resizebox{\textwidth}{!}{
  \begin{tabular}{llllllllllllllll}
\midrule
    \multicolumn{16}{c}{KV cache Compression Ratio = 1.2\%}\\
    \midrule
StreamingLLM&
19.05 & 7.15 & 20.55 & 32.40 & 26.61 & 15.70 & 16.13 & 20.42 & 17.07 & 42.00 & 73.37 & 34.52 & 54.33 & 52.65 & 30.85\\
H2O
&22.56& 13.21& 28.73& 39.24& 26.70& 19.57& \textbf{20.10}& 21.71& 20.05& 35.50& 89.04& \textbf{38.60}& 54.52& 50.81& 34.31\\
SnapKV
&20.96& 12.91& 30.83& \textbf{39.45}& \textbf{28.75}& 19.12& 19.66& 21.82& 20.45& 47.00& 88.77& 37.64& 56.41& 50.97& 35.33\\
PyramidKV
&21.33& \textbf{14.61}& 30.69& 35.62& 26.75& 18.74& 19.08& 21.81& 19.65& 59.00& 87.27& 37.01& 55.20& 50.04& 35.48\\
XKV(Ours)
&\textbf{23.0}6& 14.41&\textbf{32.94}& 38.09& 27.78& \textbf{20.16}& 19.80& \textbf{22.09}& \textbf{20.54}& \textbf{63.00}& \textbf{89.23}& 38.27& \textbf{56.63}& \textbf{53.03}& \textbf{37.07}\\
    \bottomrule
  \end{tabular}
  }
  \label{tab:table3}
\end{table*}
The suite of experiments explores the ability of memory reduction for different competitors, under a specific bound of accuracy loss. \autoref{fig:51} plots the compression ratio versus the accuracy metric. XKV demonstrates significantly superior KV cache compression capability compared to existing methods. For example, in Few-shot Learning tasks, the inference accuracy of all competitive methods begins to degrade when the KV cache compression rate reaches 51.2\%. In contrast, XKV achieves the minimum compression ratio, 25.6\%, with nearly-zero loss in inference performance. Particularly, in code completion tasks, XKV, H2O, and StreamingLLM  have prominent memory reduction performance. The inference accuracy does not degrade even at the very low compression ratio of 12.8\%. We even observe the increase trend. This is because in such tasks, the attention distribution is highly sparse. As a result, most of the important information are embedded in only a few tokens. That is naturally suitable for preserving important caching data, enabling an extremely low compression rate (even below 10\%). In general, PyramidInfer exhibits heavy accuracy degradation in all tested tasks, especially when the compression ratio is reduced to 25.6\%. This is because its static hyper-parameters cannot evolve in different applications. Notably StreamingLLM always cache the most recently generated KV-pairs and sink ones. This design cannot handle the QA tasks where the importance of tokens significantly varies and hence the important might not be cached. H2O generally has an overall prominent performance, but it still underperforms our XKV, sinc DDID is not considered.

For all datasets, XKV achieves an average minimum KV cache compression ratio of 38.4\%. This indicates that XKV can reduce the KV cache memory overhead by 61.6\% on average, which is at least 12.8\% higher than that achieved by the best competitor.

\subsection{Evaluation on Inference Accuracy}\label{sec:exp:accuracy}
We next fix the memory capacity (compression ratio) to compare the accuracy loss of competitors. Notably with the extremely low compression ratio, some K-V pairs w.r.t. tokens critical for subsequent inference are inevitably evicted. That generates very negative impact on accuracy. We thereby simulate this strict testbed. As shown in \autoref{fig:52}, we vary the compression ratio ranging from 6.4\% to 1.2\%. For the five types of long-context inference tasks, XKV consistently outperforms existing advanced competitors. SnapKV sets a fixed KV cache size for each layer, which cannot be smartly tuned on demand. Under the extreme compression scenario, many important tokens at the layer with limited cache are evicted, leading to a heavy accuracy penalty. PyramidKV enables the dynamic tuning optimization but follows a coarsen-grained pre-defined pattern. The cache size of late layers is always smaller than that of previous layers. The optimization space is thereby still very limited. \autoref{t3} provides a detailed comparison of inference accuracy across 14 datasets. Specifically, when the KV cache compression ratio is set to 1.6\%, XKV performs better than competitors on 11 out of the 14 datasets. These results demonstrate that DDID can guide XKV to perform a fine-grained tuning across layers. Then more important tokens can be preserved for better accuracy.

\subsection{Evaluation on Inference Efficiency}\label{sec:exp:runtime}
\begin{table}
 \caption{Comparison on efficiency (NaQA)}\label{t3}
  \centering
 \resizebox{0.48\textwidth}{!}{
\begin{tabular}{lrrrr}
\toprule
Methods&\makecell[l]{Total \\Runtime(s)}&\makecell[l]{Runtime of \\Single Task(s)}&\makecell[l]{KV Cache of\\Single Task(GB)}&\makecell[l]{Maximum \\Batch Size} \\
\midrule
FullKV&1436&7.18&4.00&8\\
XKV(Ours)&680&3.40&1.54&20\\
\bottomrule
\end{tabular}
  }
\end{table}
We finally analyze the inference efficiency by comparing our XKV against the underlying FullKV without eviction. We conduct the efficiency experiments on an entire dataset NarrativeQA (NaQA for short) used in the Single-Document QA inference task. \autoref{t3} summarizes the result. When the batch size is 1, FullKV takes 1436 seconds to complete all inference tasks for the entire dataset, while XKV takes 680 seconds. More specifically, XKV completes a single inference task within 3.4 seconds, which is still faster than FullKV (7.18 seconds). The computational efficiency is improved by 2.1$\times$. We next give our explanation. Recall that XKV can evict data at the very beginning of an inference task. Such eviction reduces the memory consumption of pre-filling from 4GB to 1.54GB. That not only reduces the number of cached data, but also the associated workloads in subsequent generation procedures. The inference speed of a single task is thereby fast. Furthermore, the memory reduction of a single task enables a large batch size (from 8 to 20), that significantly improves the utilization of the GPU compute power. Due to the two factors mentioned above, compared with FullKV, our XVK can increase the throughput by up to 5.2$\times$.

\section{Conclusion}
In this work, we proposed XKV, a novel approach to achieving personalized KV cache reduction during inference by leveraging DDID across the hierarchical structure of LLMs. Through detailed experimental investigations and theoretical analyses, we modeled KV cache reduction as an optimization problem guided by DDID. To solve this, we introduced a new inference framework incorporating adaptive allocation and dynamic eviction algorithms, along with a sampling-based strategy to mitigate computational overhead. Extensive experiments validate the effectiveness of XKV.

\bibliographystyle{IEEEtran}
\bibliography{references}

\end{document}